\documentclass[letterpaper, 10 pt, conference]{ieeeconf} 
\IEEEoverridecommandlockouts                             
\usepackage{color}

\usepackage{graphics} 
\usepackage{epsfig} 
\usepackage{mathptmx} 
\usepackage{times} 
\usepackage{amsmath} 
\usepackage{amssymb}  
\usepackage{subfigure}
\usepackage{multirow}
\usepackage{graphicx, epstopdf}
\usepackage{xspace,epsfig,url}
\usepackage[noadjust]{cite}
\usepackage{enumerate}
\usepackage{float}
\usepackage{epsf}
\usepackage{psfrag}
\usepackage{verbatim}
\usepackage[all]{xy}
\usepackage{color}
\usepackage[table]{xcolor}
\usepackage{bm}
\usepackage{cases}
\usepackage{hyperref}
\usepackage[ruled,vlined,linesnumbered]{algorithm2e} 

\title{\LARGE \bf Dynamically Reconfigurable Discrete Distributed Stiffness\\ for Inflated Beam Robots}

\author{Brian H. Do, Valory Banashek, and Allison M. Okamura
\thanks{The authors are with the Dept. of Mechanical Engineering, Stanford University, Stanford, CA 94305, USA. Email: \{brianhdo, valoryb, aokamura\}@stanford.edu} \thanks{This work is supported in part by the National Science Foundation Graduate Research Fellowship Program and by Toyota Research Institute (TRI). TRI provided funds to assist the authors with their research, but this article solely reflects the opinions and conclusions of its authors and not TRI or any other Toyota entity.}%
}

\begin{document}

\maketitle
\thispagestyle{empty}
\pagestyle{empty}

\begin{abstract} 


Inflated continuum robots are promising for a variety of navigation tasks, but controlling their motion with a small number of actuators is challenging. These inflated beam robots tend to buckle under compressive loads, producing extremely tight local curvature at difficult-to-control buckle point locations. In this paper, we present an inflated beam robot that uses distributed stiffness changing sections enabled by positive pressure layer jamming to control or prevent buckling. Passive valves are actuated by an electromagnet carried by an electromechanical device that travels inside the main inflated beam robot body. The valves themselves require no external connections or wiring, allowing the distributed stiffness control to be scaled to long beam lengths. Multiple layer jamming elements are stiffened simultaneously to achieve global stiffening, allowing the robot to support greater cantilevered loads and longer unsupported lengths. Local stiffening, achieved by leaving certain layer jamming elements unstiffened, allows the robot to produce ``virtual joints" that dynamically change the robot kinematics. Implementing these stiffening strategies is compatible with growth through tip eversion and tendon-steering, and enables a number of new capabilities for inflated beam robots and tip-everting robots. 

\end{abstract}

\section{INTRODUCTION} 




Due to their inherent compliance, soft robots possess an infinite number of degrees of freedom (DOFs), which is often advantageous for navigation in cluttered or unstructured environments. It allows them to passively adapt to their environment \cite{brown2010universal,rus2015design}, a characteristic often described as ``embodied intelligence". 
Their compliance can also enable safer human-robot interactions.
However, soft robots typically control only a few of these DOFs, limiting their effectiveness for certain tasks. Active control allows for richer, more diverse interactions with the environment. Here we focus on active control of stiffness. 
Because of the unique advantages inherent to both compliant and rigid robots, soft robots can greatly benefit from stiffness control \cite{cianchetti2013stiff,manti2016stiffening,Wall2015,Amend2012,Jiang2012}. 

Achieving such stiffness control in a soft robot typically requires a large number of controllable DOFs, which brings about its own challenges. Conventionally, the greater the number of active DOFs, the more actuators are needed. In soft robots, this can be infeasible due to practical design limitations. However, the control of DOFs may be simplified by using an activating mechanism to modify actuator outputs, rather than using more actuators \cite{steltz2009jsel}. An activator is a mechanism that does no work on its own, but serves to modify and modulate an actuator's output.


In this paper, we introduce the concepts behind using discrete distributed stiffness to enhance the capabilities of inflated beam robots and present an implementation of this idea via positive-pressure layer jamming, which uses a series of custom bi-state passive valves to separate actuation and activation. This is incorporated into a type of inflated beam robot capable of tip extension (``growth'') through pressure-driven eversion~\cite{HawkesScienceRobotics2017}. We demonstrate the ability of this robot to control and actively modify its morphology into high-DOF shapes with only two actuators, prevent buckling under transverse loading through increased stiffness, and grow via tip eversion with these added features. 

\section{BACKGROUND} 
Here we discuss previous actuation approaches in inflatable beam robots, methods to modulate the outputs of soft robots, and ideas on how to simplify control of those outputs.

\begin{figure*}[th!]
      \centering
      \includegraphics[width=\textwidth]{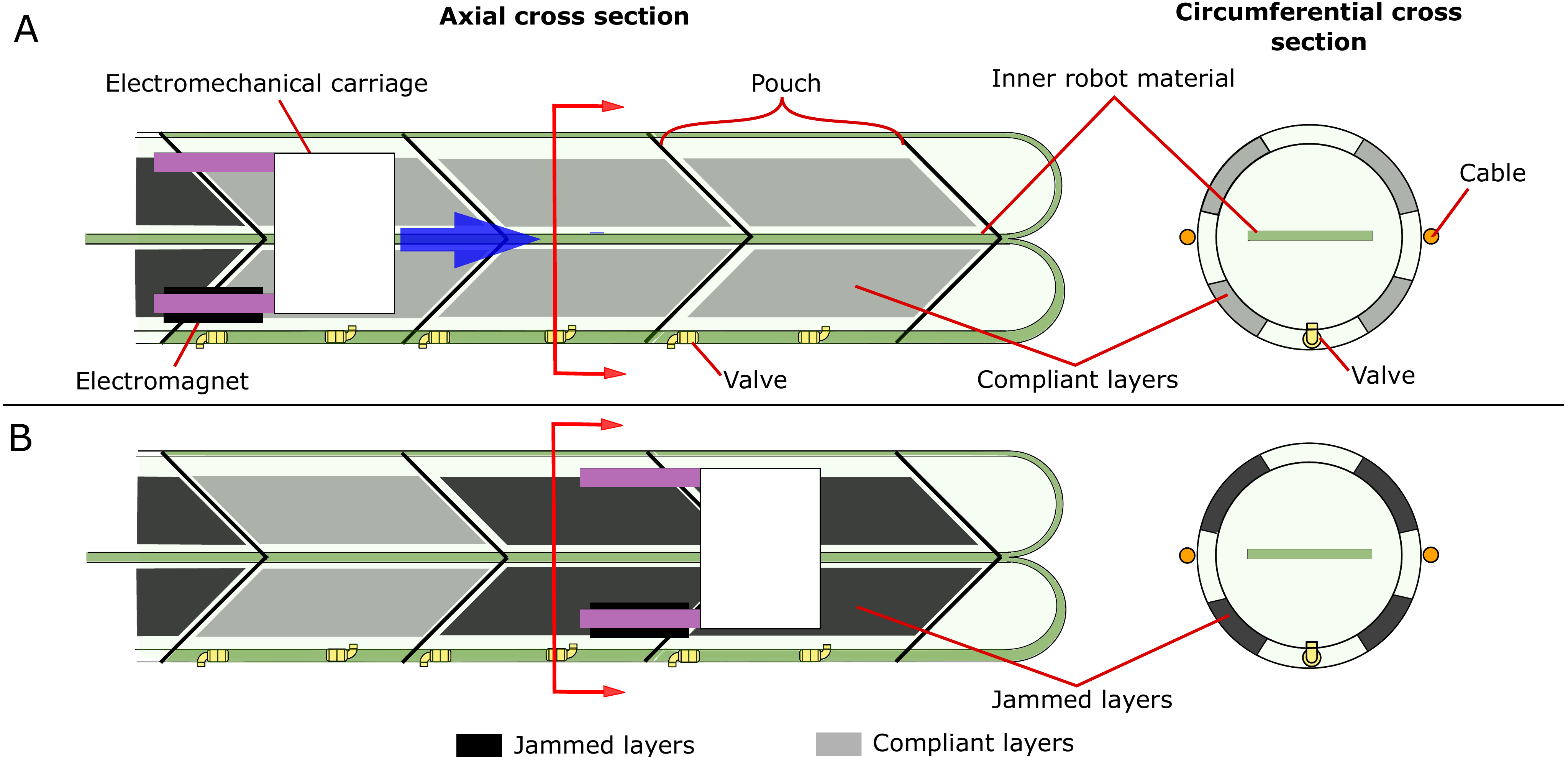}
      \caption{Cross sections of the distributed discrete stiffness robot. (A) Side axial and circumferential cross sections of the robot. The robot is composed of a series of stiffening pouches along the main beam axis. Each pouch wraps completely around the robot circumferentially and contains four parallelogram shaped layer sections. Each pouch also contains two valves whose states can be selectively switched using an electromagnet carried by an electromechanical carriage. This carriage drives along the inside of the robot on the inner robot material. The entire robot can be steered via two cables that run along the robot's left and right sides. (B) As the carriage travels the length of the robot, pouches can selectively be made stiff or compliant via layer jamming.}
      \label{DesignConcept}
      \vspace{-0.33cm}
  \end{figure*} 

\subsection{Inflated beam robots}
Most continuum manipulators feature a flexible but rigid backbone which acts as a support structure. In contrast, inflated beam robots are characterized by the use of internal gas pressure as the supporting element \cite{comer1963deflections}. Because of this, inflated beam robots can provide inherently safe human-robot interaction due to their compliance and low inertia \cite{sanan2011physical}, and can have long, thin aspect ratios, making them well suited for tasks such as inspection \cite{allen2003robotic, voisembert2013design, perrot2010long}. However, the dexterity of inflated beam robots comes at the cost of complex mechanical design. Past inflatable robotic arms have relied either on actuators located at the joints \cite{koren1991inflatable} or run cables from each joint down the length to a base \cite{voisembert2013design, perrot2010long}. 

A special class of inflated beam robots are tip-extending, pneumatically everting robots. These robots are composed of a thin-walled, compliant, inextensible material which is initially inverted. Application of pressure then causes the robot to lengthen at the tip as it draws new material through itself \cite{HawkesScienceRobotics2017, mishima2003development, tsukagoshi2011tip}. These tip-extending robots are capable of navigating their environment through growth and have been steered by (1) preforming robots with desired shapes, (2) using pneumatic actuators to achieve curvature along the robot length \cite{blumenschein2018helical}, or (3) interacting with environmental obstacles and buckling \cite{greer2018obstacle}.  

For any inflated beam robot, there exist critical axial and transverse forces above which it will buckle \cite{coad2019retraction, godaba2019payload}. Typically, such buckling is to be avoided, as it may distort the robot from a desired configuration. 
In contrast, for the distributed discrete stiffness robot presented in this work, we intentionally allow buckling at desired, changeable locations by configuring patterns of stiff and compliant pouches, and utilizing that buckling to produce desired shapes and turns. 

\subsection{Modulating Actuator Output}
Like other continuum robots, inflated beam robots have few controllable DOFs due to the large number of actuators required to realize such control. One approach to enable the design of such robots with high active DOFs is separating actuation and activation \cite{steltz2009jsel}. Actuators enable the robot to do work on the environment, while activators do no work on their own. By using a large number of simple activators that modify and modulate the the output of a few actuators, robots can require substantially fewer complex actuators. 



Jamming is an approach to modulating the forces exerted by soft robots; here we use it as a technique for activation. Jamming involves generating a normal force on layers or particles that compresses them together, increasing the friction force between them and thus the total stiffness \cite{kim2013novel}. Jamming can be accomplished by placing thin layers or particles of a material in an airtight membrane and then generating a negative pressure gradient across that membrane. Previous jamming work for soft robots produced tunable stiffness structures that transition from being compliant to rigid, allowing the robots to retain their shape during physical interaction \cite{Follmer2012,Wall2015,Amend2012,Jiang2012}. In contrast, layer jamming is used here to create ``virtual joints" to enable greater shape change and maneuverability. 



Because they require a negative pressure gradient, most jamming devices rely on a vacuum source \cite{manti2016stiffening,cianchetti2013stiff,cheng2012design}. However, a vacuum source is not strictly necessary. By operating a jamming device in an environment above atmospheric pressure, the atmosphere can be used as a low pressure source. This can be done with inflated beam robots, since they, by definition, involve inflating beams above atmospheric pressure. We implement this ``positive-pressure layer jamming" so our robot has no internal pressure lines and requires no additional auxiliary pressure source.


Typically, a robot containing $n$ layer jamming cells would require at least $n$ vacuum lines, $n$ valves, and $2n$ wires to regulate flow to/from the atmosphere and a vacuum source. This can make it infeasible to create a layer jamming robot with a high number of jamming cells. 
This remains an issue even when separating actuators and activators \cite{steltz2009jsel}. 
Other stiffening mechanisms, such as shape memory polymers \cite{Firouzeh2017} and phase changing alloys \cite{Alambeigi2016}, have the same challenges. 




\section{Design Concept} 


 We present a robot with dynamically reconfigurable distributed discrete stiffness along its length which builds on the actuation/activation paradigm. Along the length of the robot are a number of pouches containing layers. By modifying the pressure inside these pouches, the layers can be stiffened via layer jamming. This design extends the activation/actuation paradigm with the design of bi-state valves, which passively prevent airflow in the presence of a pressure gradient and offboard their activation to an electromagnet, which is driven by an internal electromechanical carriage.

Fig.~\ref{DesignConcept} shows a side view diagram of the robot. The inflated beam body is made from a flexible, inextensible material to allow for apical extension via pneumatic eversion. As such, there is material that runs inside of the robot body. However, this dynamically reconfigurable distributed stiffness concept can also apply to non-everting beam robots.

An internal electromechanical carriage containing an electromagnet drives inside of the main robot body to switch the state of small, simple bi-state passive valves. Because the mechanism that switches the valve states is offboarded onto the carriage, the valves can be switched without explicitly running wires to each valve, allowing them to be made very simply and inexpensively. This makes it possible to mass produce these valves for an extremely long robot. The only set of wires needed for this robot are the four wires needed to control the motors of the internal carriage and the two for the electromagnet; this remains constant regardless of the number of valves or jamming pouches. 

The carriage uses the inner robot material as a rail to drive along the length of the robot body. However, other methods such as driving off the outer body material or pushing an electromagnet through an airtight throughhole in the robot body would also work. 

By unjamming pouches, locally compliant regions can be created. The application of an axial force via a cable then can allow for buckling at this point, producing a ``virtual joint." Rejamming a pouch in a region that has been bent locks the joint angle, and unjamming a new pouch allows a new joint to be created. With this, two motor-pulled cables can produce a high number of possible shapes.

This decreased design and fabrication complexity comes at the cost of response time. While the pouches themselves have a fast state transition from jammed to unjammed, they must be addressed sequentially, and the time required to switch their state now includes the travel time of the internal carriage as it drives to the pouch location. For situations in which fast response times are not needed however, this trade-off results in significant benefits. Furthermore, this issue is mitigated during growth for tip-extending inflated beam robots, as the carriage can remain at the robot tip if desired and can jam pouches as soon as they are everted. 

\section{DESIGN AND IMPLEMENTATION} 

\subsection{Soft robot body with layer jamming pouches}
The robot body was made from a low density polyethylene plastic tube. The jamming layers were laser cut from polyester fabric and secured to the main robot body using double-sided tape. We placed the layers along the length of the plastic tube and then heat sealed pouches around the layers with an impulse sealer to form airtight seals. After completion of all pouches, the plastic is inverted so that the pouches are now internal to the main robot body. 

Prior to buckling, inflated beams experience wrinkling. As greater force is applied to the beam, these wrinkles propagate in a ring around the beam until it reaches a critical point, after which it buckles \cite{liu2016interactive}. As a result of this, chevron-shaped layers were used rather than rectangular layers to prevent undesired buckling by leaving uninterrupted lines around the robot body's circumference where it is not reinforced by jammed layers. The chevron pouches fit into each other as shown in Fig.~\ref{DesignConcept}.

\begin{figure}
      \centering
      \includegraphics[width=0.9\columnwidth]{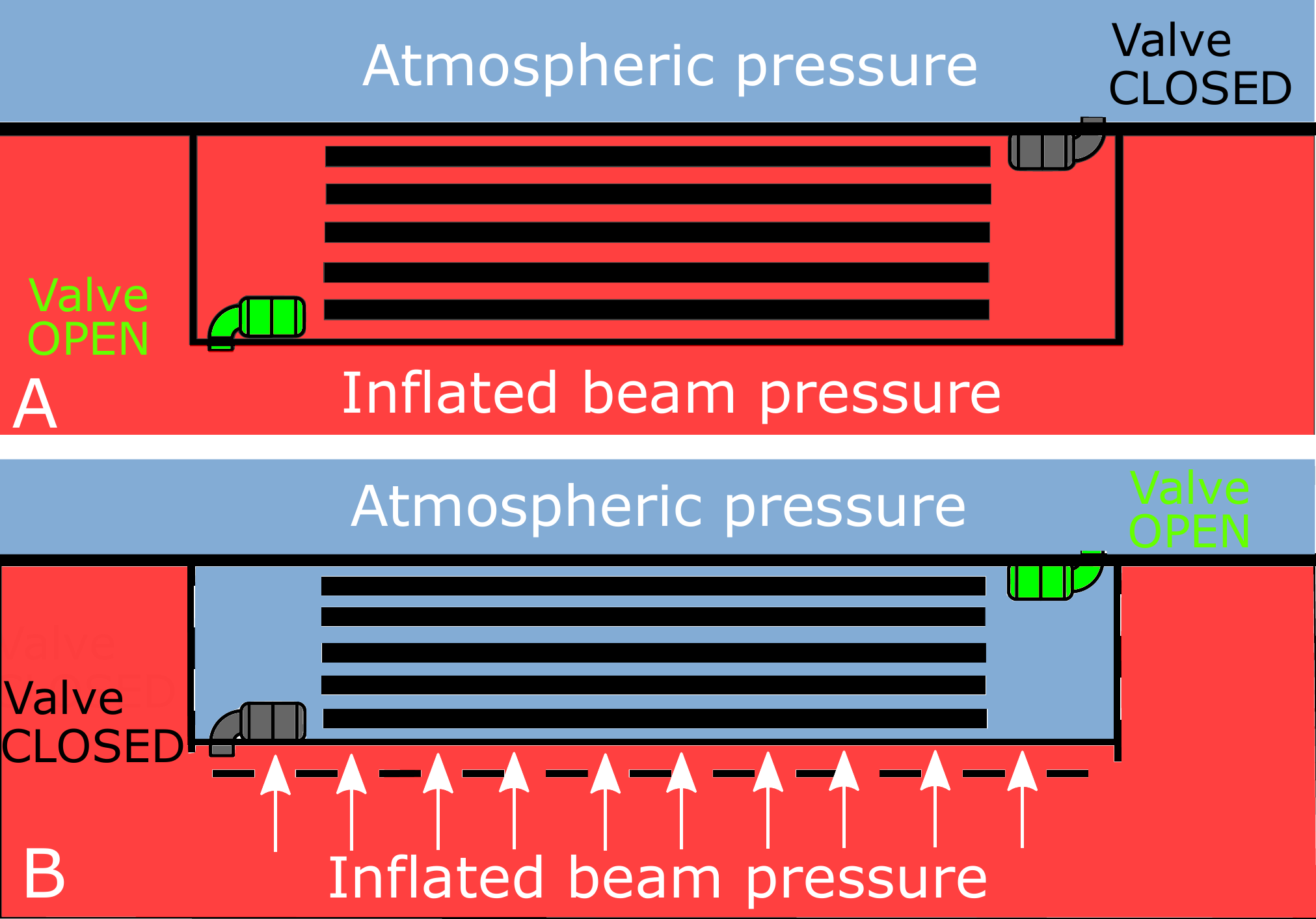}
      \caption{Principle of operation for layer jamming pouches. (A) A series of pouches line the inside of the inflated beam robot body. The pouch are set to either atmospheric pressure or to the higher inflated beam pressure. When the innermost valve leading from the robot body to the pouch is open, the pouch and beam pressures equalize. In this state, the layers are compliant. (B) When the innermost valve is closed and the valve connecting the pouch to the atmosphere is opened, the pouch depressurizes to atmospheric pressure. As pouch pressure drops, the inflated beam pressure produces a normal force on the pouch, compressing it from its previous inflated state (as represented by dotted lines) and jamming the layers. Now, the layers are stiffened.}
      \label{PouchOperation}
  \end{figure}


Fig.~\ref{PouchOperation} illustrates how a pouch operates. To discretely switch the layer state from maximally jammed to unjammed, two passive valves are used per pouch. One valve controls airflow from the main robot body into the pouch. The second controls airflow from the pouch into the atmosphere. To set the pouch pressure to either atmospheric pressure or the inflated beam pressure, the corresponding valve connecting the valve to that pressure source is opened and the other closed. Hot glue secures the valves to the pouches.

At the base of the inflated beam robot are two motorized spools which pull two cables that run along the robot body's length. Short PTFE stoppers are attached along the left and right sides of the robot and provide a guide for the cables.

\subsection{Bi-state passive valves} 

\begin{figure}
      \centering
      \includegraphics[width=0.9\columnwidth]{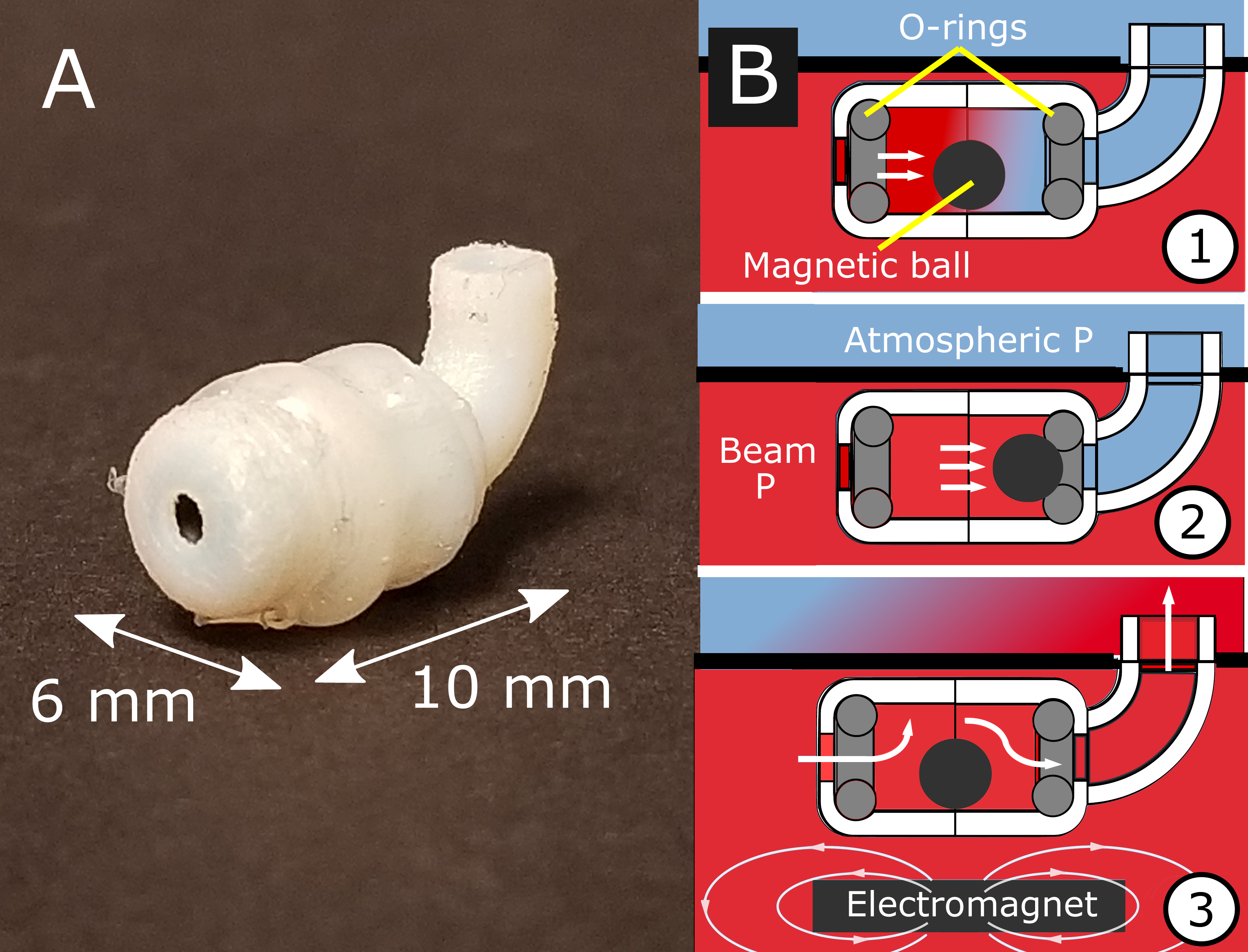}
      \caption{Valve design and operation. (A) Photograph of one 3-D printed valve used in the inflated beam robot. (B) Sequence illustrating the valve operation. The valve has an inner hollow channel containing two O-rings and a magnetic ball. (1) If the valve is initially opened and a pressure gradient exists, air can flow through the valve. If not, the ball will just sit in the valve. (2) This flow pushes the ball into one of the O-rings, forming a seal and preventing further flow. (3) The valve can be re-opened by bringing an electromagnet near the valve. This pulls the magnetic ball away from the O-ring.}
      \label{ValveOperation}
  \end{figure}

Fig.~\ref{ValveOperation}A shows a picture of one of the valves used for the pouches. Fig.~\ref{ValveOperation}B shows a cross section of a valve. The valves were 3D printed in halves using a Stratasys Objet30 Pro, and an O-ring was inserted into a groove in each half. A 2.5~mm diameter neodymium sphere (Speks) was placed in the valve assembly, and the two halves were secured together and to the plastic walls of the inflated beam robot using hot glue.

Each valve is unpowered and externally actuated, with the only moving part being the internal magnetic ball. Fig. \ref{ValveOperation}B illustrates how the valve operates. In the presence of a pressure gradient, the air through the valve pushes the internal magnetic ball into an O-ring, producing a seal and thus preventing airflow. When an external electromagnet is brought in proximity to the valve, the magnetic ball is attracted to it and moves away from the O-ring, allowing airflow. In the absence of a pressure gradient, the magnetic ball simply sits in the middle of the valve and does not form a seal with the O-ring.

The mechanical simplicity of the valves allows them to be made quite small. The main body of each valve is 10~mm long and 6~mm in diameter, with a small tube that protrudes out from one side an additional 3.5~mm in length and 2~mm in height. Inside of the main body is a cylindrical cavity 4.5~mm in diameter along with two small grooves for the O-rings to rest in. One completed valve weighs about 0.4~g and costs USD 0.19 in raw materials. This small size and low cost allows for many valves to be embedded throughout the length of an inflated beam. 

If conventional valves were used, the manufacturing complexity of the beam skin would be greatly increased due to the need to wire each valve and integrate these wires into the skin. This becomes challenging for extremely long inflated beam robots, as the number of wires is $2n$, where $n$ is the number of valves. Implementing this with growth via tip eversion would be particularly difficult because the wall material transitions from being on the inside of the robot to the outside as the robot lengthens.  

\subsection{Electromechanical carriage}

\begin{figure}
      \centering
      \includegraphics[width=\columnwidth]{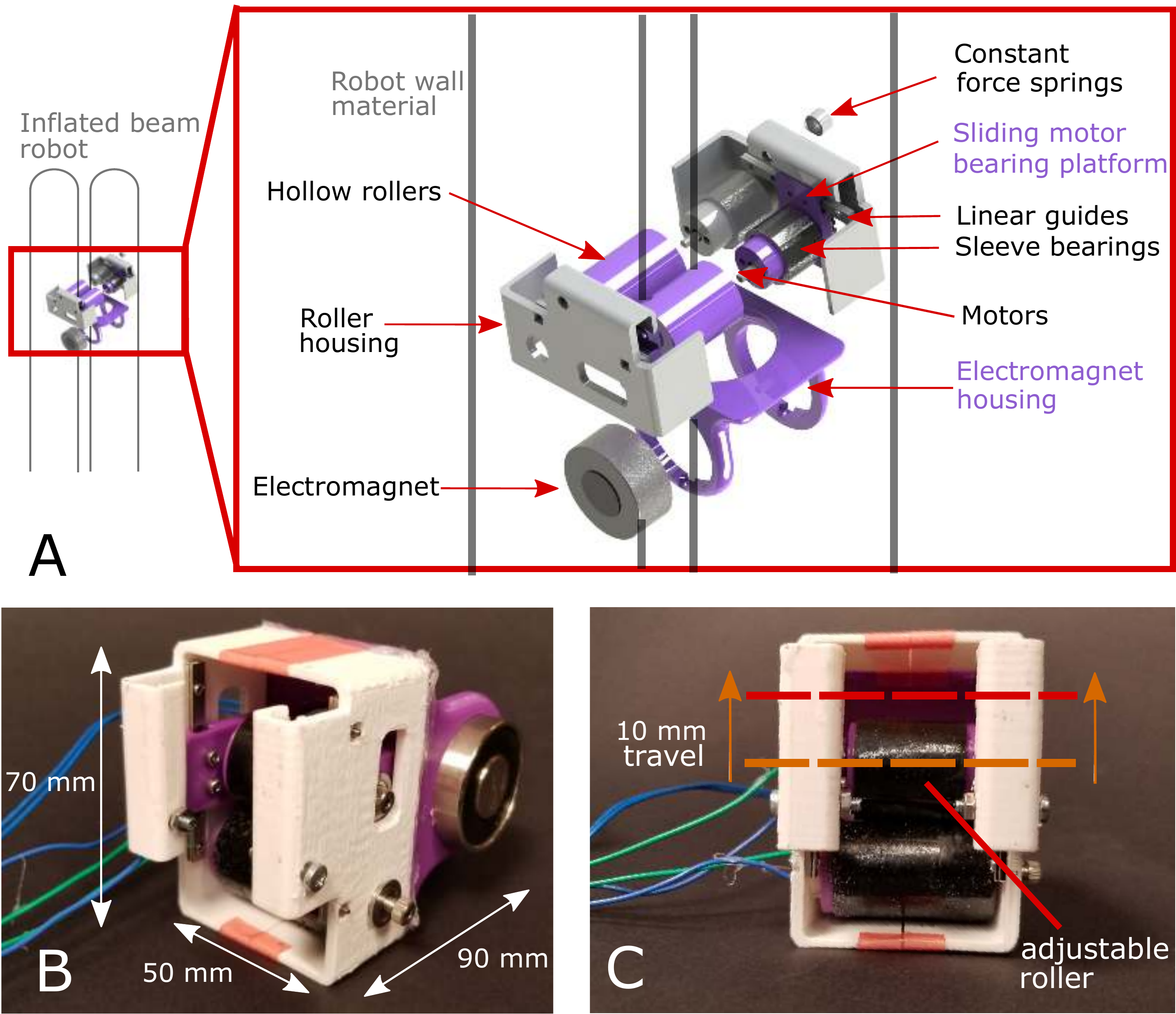}
      \caption{Implementation of the electromechanical carriage to carry the electromagnet. (A) Exploded view of the carriage. The device contains a roller mounted onto a linear guide. This allows the roller to translate and roll over layers of jamming material and valves while the distance between the rollers passively adjusts. (B) The sliding motor bearing platform and side groove to allow for translation of wires and bearings can be seen in this photograph as well as overall device dimensions. (C) Front view of the device. The adjustable width roller can move from its default position (orange dotted line) up to an additional 10 mm (red dotted line).}
      \label{DeviceCAD}
  \end{figure}

\begin{figure*}[t!]
      \includegraphics[width=2\columnwidth]{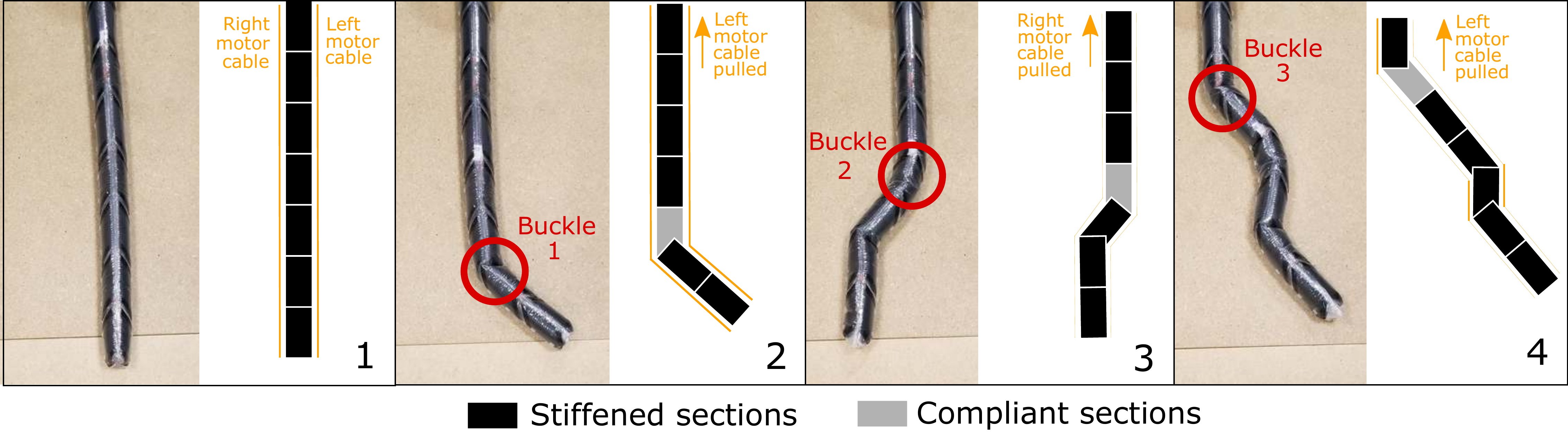}
      \caption{Sequential buckling of the robot. (1) Initially, all sections in the robot are stiffened. (2) By unjamming the layers in one pouch and then pulling the left motor cable, Buckle 1 is formed. (3) This is then rejammed and a (new pouch is unstiffened. Pulling on the right motor cable produces Buckle 2. Buckle 1 still remains and the shape produced by that buckle remains fully intact. (4) Continuing with this process, the previously buckled pouch is restiffened, a new pouch is unstiffened, and a buckle is formed by pulling the left motor cable. The shapes produced by Buckles 1 and 2 remain.}
      \label{SequentialBuckling}
      \vspace{-0.33cm}
  \end{figure*}

Fig. \ref{DeviceCAD} shows the electromechanical carriage. Previous work developed a device that traverses inside of and along the length of an inflated beam robot \cite{coad2019retraction}; our carriage builds on that design. The carriage consists of two main parts: a housing for adjustable width rollers and an electromagnet. The housing and rollers are both 3-D printed. Each roller slides over a motor housing and is rigidly connected to the motor shafts using set screws. The rollers roll on needle roller bearings mounted on the motor housing at one end and connect to the other half of the roller housing via ball bearings held in place by shoulder screws. Each roller is wrapped in Dycem non-slip material (Dycem Corporation) to prevent slipping with the inner robot material. Hot glue secures the two halves of the roller housing.

One of the rollers and its corresponding motor and bearings is mounted on two sliding bearing platforms attached to linear guides housed in the sides of the device. These platforms are tensioned with constant force linear springs to ensure solid contact between the rollers and the inner robot material. The distance between the rollers passively adjusts as the carriage moves over the layers and the valves as it drives along the inner robot material. Two small screws are used as a mechanical stop to keep a minimum roller distance. 
A 3-D printed electromagnet housing for a 12~V electromagnet (uxcell) is attached to the back of the motor housing. The wires from the motorized rollers and the electromagnet run all the way through the soft robot body from the carriage to the base.
The electromagnet constrains the carriage's design. The device measures approximately 70~mm by 50~mm by 90~mm, with much of the length due to the electromagnet holder. The device also weighs 202~g, with the electromagnet responsible for about half of that weight. 

One major trade-off with using a carriage is the speed with which a set of valves can be actuated. Because the carriage drives along the robot body, valves are only sequentially activated, and the time required to switch valve states now includes the time for the carriage to travel to the valves. For inflated beam robots which navigate through pressure-driven eversion, this is not an issue, as previously grown segments remain stationary relative to the environment. As such, the carriage only needs to drive at the tip, and the inflated beam robot can continue to navigate its environment in a responsive manner. Furthermore, multiple carriages could be deployed simultaneously, decreasing the distance that each carriage travels but increasing the number of wires required. 
  
\section{DEMONSTRATIONS AND EXPERIMENTS} 
\subsection{Active shape changing and reconfiguration} 

One new capability made possible by reconfigurable distributed stiffness is producing a variety of shapes with non-continuous curvature along the robot length using just two pull-string actuators. Fig.~\ref{SequentialBuckling} demonstrates the main process by which the robot can be actively reconfigured into a new shape. An 8.6~cm diameter, 1.2~m long discrete stiffness robot containing 8 stiffening pouches was used. The robot body was actuated in one DOF (left/right) by two cables which run along the robot and connect to motorized spools at the robot base.

Initially, the robot is completely stiffened as all the layers in the pouches are jammed. The electromechanical carriage then drives down the length of the robot and unjams the layers in a specified pouch by activating the electromagnet over the appropriate valve. Pulling on one of the cables now produces a buckle at that pouch location as the critical buckling force is higher for all other stiffened sections compared to the critical buckling force at this pouch. That buckling point now acts as a ``virtual joint", and the subsequent robot sections can pivot about that buckling point, producing a turn in the robot. The carriage then drives back towards the base, and as it does so, jams the layers in the aforementioned pouch. This re-stiffens the pouch, preserving the robot shape at that ``virtual joint" throughout subsequent bends. 

\begin{figure} 
      \centering
      \includegraphics[width=0.8\columnwidth]{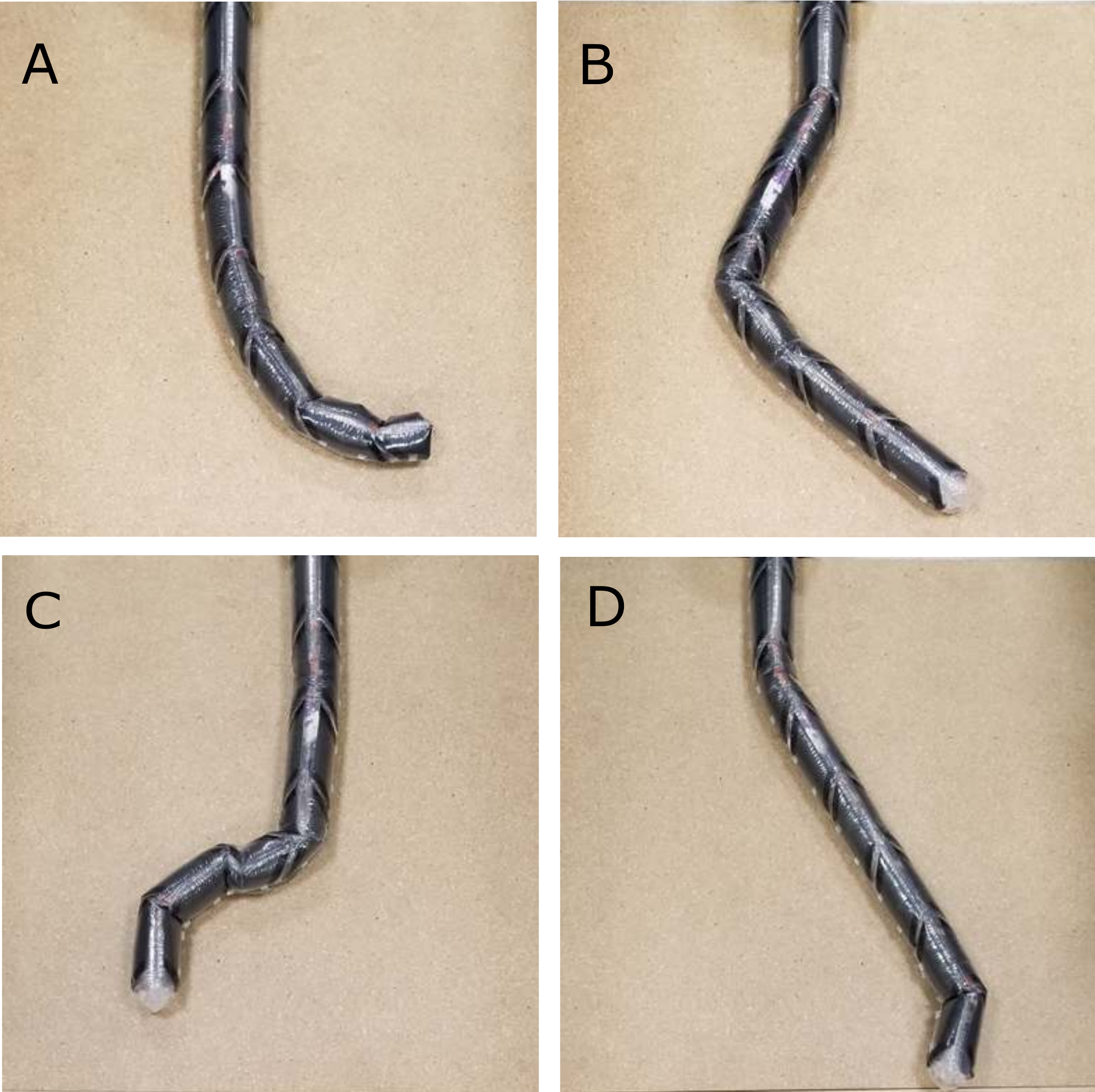}
      \caption{The same distributed stiffness inflated beam robot can be actuated into a variety of different shapes by pulling on cables on the sides of the robot. (A) By jamming all pouches, the robot can be made very stiff, allowing it to be pulled into a curved segment without buckling at the base. (B-D) By selectively stiffening certain sections, local buckling can occur, resulting in non-continuous curvature segments. Segments can be bent either left or right by pulling on the appropriate cable. This allows for shapes of the robot with curvatures in two different directions, which is not possible without direct control of stiffness.} 
      \label{ActuatedShapes} 
  \end{figure} 

Furthermore, we can use the same discrete stiffness robot to produce a wide variety of shapes. Fig.~\ref{ActuatedShapes} shows this capability, with the robot producing four distinct shapes. This is done by resetting the robot by unjamming previously jammed layers, allowing the robot to straighten, and repeating the process discussed in Fig.~\ref{SequentialBuckling}.

More global curvature can be produced in addition to local non-continuous curvature. Fig.~\ref{ActuatedShapes}A demonstrates this. By stiffening all pouches, the entire robot body can be made very stiff. Because of this inherent stiffness, the robot body does not buckle at higher applied cable forces, allowing the tip to be pulled into a constant curvature section.

In general, each pouch along the robot body can buckle to a variety of arbitrary angles in either the left or right direction. Previous buckling shapes can be preserved and chained together to achieve complex bending shapes. For example, this allows for shapes of the robot with curvatures in two different directions, which is not possible without direct control of stiffness. This represents a new capability for general inflated beam robots: previously, producing such local turning along a beam's length would have required multiple actuators -- one for each turning point. 


\subsection{Increased resistance to cantilevered loads} 

\begin{figure}
      \centering
      \includegraphics[width=0.75\columnwidth]{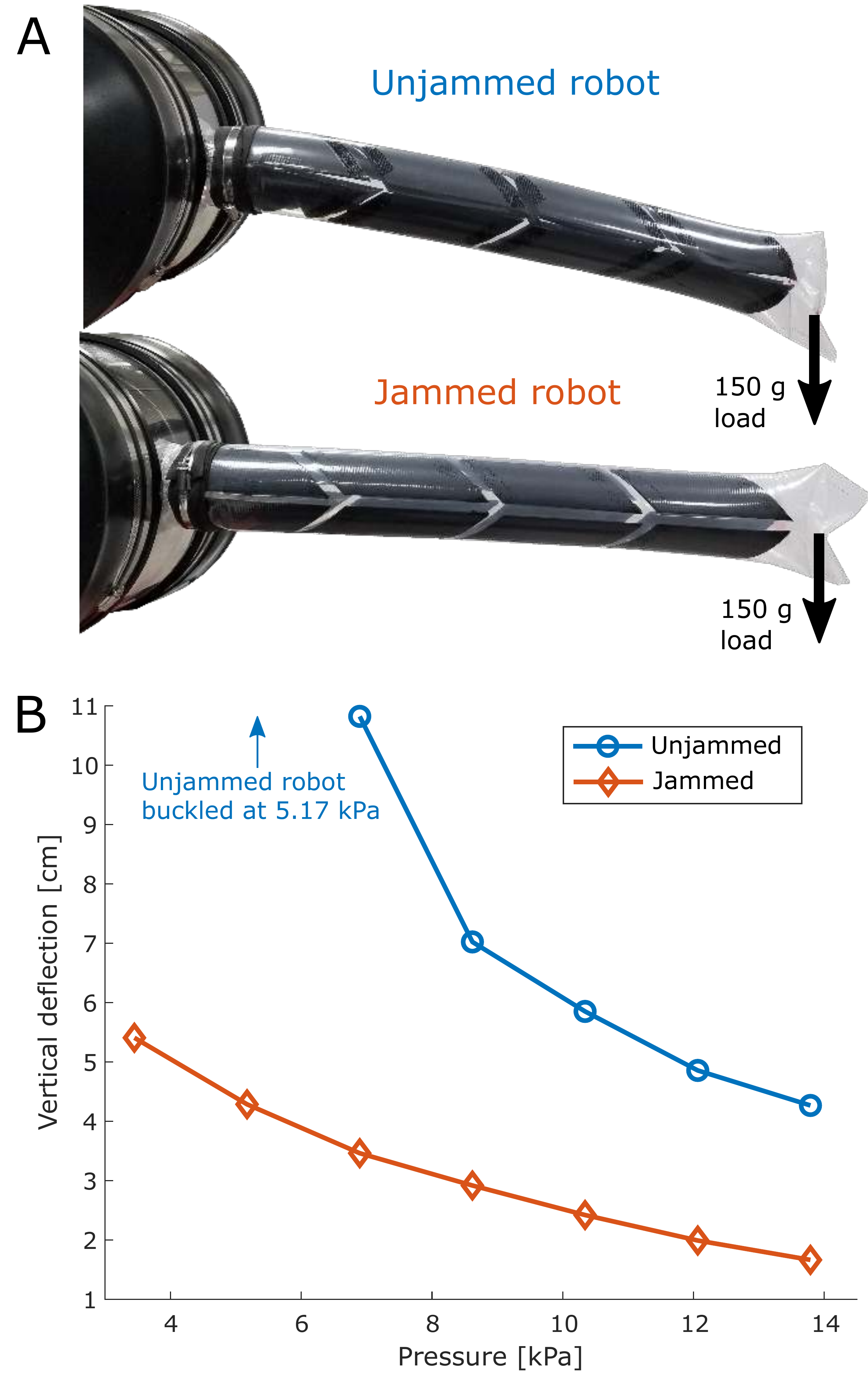}
      \caption{Experimental results for transverse loading of unjammed and jammed inflated beam robots. A) Unjammed and jammed inflated beam robots at 60~cm in length were applied a transverse load of 150~g at their tip and the corresponding tip deflection was recorded for pressures from 3.4~kPa to 13.8~kPa. Shown is a photo of an unjammed and jammed robot at 6.9~kPa under a 150 g load. There is noticeably more curvature with the unjammed robot than with the jammed robot. (B) The deflection for the jammed robot is less than that of the unjammed robot at all corresponding pressures, and it is able to resist buckling at lower pressures.}
      \label{DeflectionTest}
      \vspace{-0.15cm}
  \end{figure}

By stiffening all sections along the length of the inflated beam robot, the global stiffness of the robot can be increased. To quantify the stiffness change between completely unjammed and jammed robots, we conducted transverse loading experiments with a 150~g transverse load applied by hanging weights from a string at the tip of a 8.6~cm diameter, 60~cm long cantilevered inflated beam robot composed of four stiffening pouches. The robots were pressurized from 3.4~kPa (0.5~psi) to 13.8~kPa (2.0~psi) in 1.7~kPa (0.25~psi) increments and pressure was controlled using a pressure regulator (Proportion Air). The tip deflection was measured with respect to a marker placed at the tip. Fig. \ref{DeflectionTest} shows experimental results from these tests. 

Fig. \ref{DeflectionTest}(B) shows that even at relatively low pressures, there is a major difference in stiffness between unjammed and jammed robots. For example, at 5.2~kPa, the unjammed robot buckled at its base, whereas the jammed robot experienced less tip deflection than the unjammed robot did at 13.8~kPa.



This stiffening capability could allow the robot to cross longer unsupported gaps between surfaces or to deliver higher payloads without significantly bending or buckling. We can modulate the robot to be soft and compliant for certain situations and then stiffen the robot to allow for greater transfer to or resistance from environmental forces.

Tip-extending robots can passively interact with environmental obstacles and buckle in order to navigate. With the addition of distributed stiffness control, these interactions could be more finely modulated, allowing more precise navigation and greater potential for movement.
  
\subsection{Compatibility with growth via pressure-driven eversion}
\begin{figure}
      \centering
      \includegraphics[width=0.9\columnwidth]{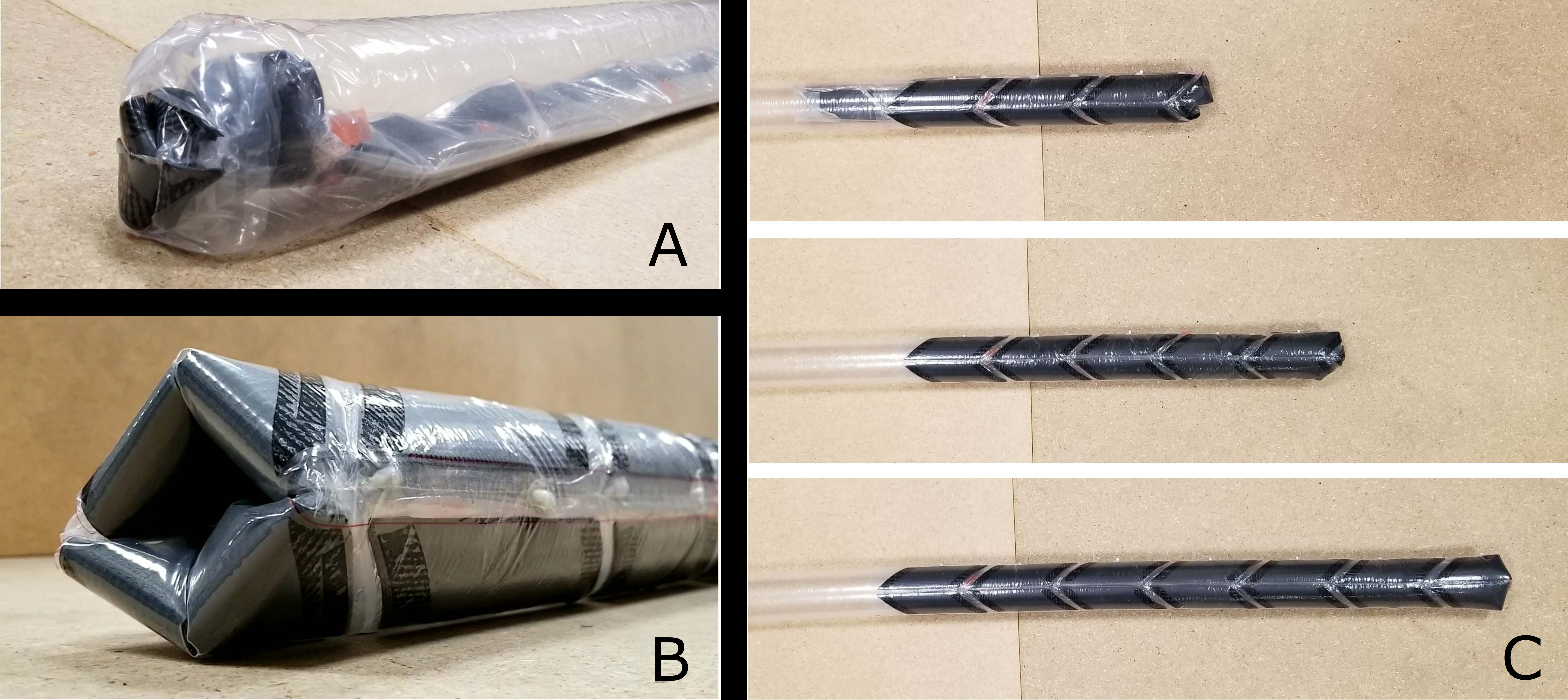}
      \caption{Compatibility of distributed stiffness with growth. (A) In this robot with a completely clear body, the inner material can be seen lying flat. (B) Image of the tip of the robot as it is everting. The tip forms a square cross section due to the four layers segments. (C) Growth of the robot over time.} 
      \label{Growth}
      \vspace{-0.15cm}
  \end{figure}
  
Growth via pressure-driven eversion introduces two main design constraints: (1) minimize the wall cross-sectional area and (2) retain flexibility for the high curvature experienced at the tip; otherwise, the robot will be unable to grow. As a result, the layers in the inflated beam robot are divided circumferentially into four sections. This allows the layers to neatly fold onto themselves and be drawn through the inside of the inflated beam as seen in Fig. \ref{Growth}A; this  also allows the carriage to drive along the robot length. 

This layer division also aids in eversion at the tip. Fig.~\ref{Growth}B shows how at the tip of the inflated beam robot, the eversion produces a square cross section. Fig.~\ref{Growth}C demonstrates the growth of such a robot over time. The material is pulled through the body and added at the tip, while previously grown sections remain stationary relative to the environment. 

This compatibility with growth via pressure-driven eversion further extends the utility of the discrete stiffening robot. Turns can dynamically be made and the robot grown without concern for buckling at the base of the robot, which represents a new capability for this class of robots.


\section{CONCLUSION AND FUTURE WORK}
We presented the concept for and the design of a novel dynamically reconfigurable distributed stiffness robot implemented using positive-pressure layer jamming. By separating actuation and activation and using simple valves that are externally switched via an electromechanical carriage, an easily scalable design was presented. We demonstrated several new capabilities for both inflated beam robots in general and tip-everting robots in particular, such as the ability to perform active shape changing and kinematic reconfiguration using the same robot with only two actuators. 

In the future, we will develop an analytical model to elucidate how stiffening via layer jamming affects the stiffness of inflated beam robots in bending and create a design tool. We will also develop a configuration planner that builds on this model to produce sequences of stiff and compliant pouches to realize arbitrary desired final shapes. As part of this, we will extend the steering work presented in this paper from 2-D to 3-D. 


\section*{ACKNOWLEDGMENT}
The authors would like to thank Margaret M. Coad for assistance in the design of the electromechanical carriage.


\newpage
\bibliographystyle{IEEEtran}
\bibliography{library}

\begin{thebibliography}{10}
\providecommand{\url}[1]{#1}
\csname url@samestyle\endcsname
\providecommand{\newblock}{\relax}
\providecommand{\bibinfo}[2]{#2}
\providecommand{\BIBentrySTDinterwordspacing}{\spaceskip=0pt\relax}
\providecommand{\BIBentryALTinterwordstretchfactor}{4}
\providecommand{\BIBentryALTinterwordspacing}{\spaceskip=\fontdimen2\font plus
\BIBentryALTinterwordstretchfactor\fontdimen3\font minus
  \fontdimen4\font\relax}
\providecommand{\BIBforeignlanguage}[2]{{%
\expandafter\ifx\csname l@#1\endcsname\relax
\typeout{** WARNING: IEEEtran.bst: No hyphenation pattern has been}%
\typeout{** loaded for the language `#1'. Using the pattern for}%
\typeout{** the default language instead.}%
\else
\language=\csname l@#1\endcsname
\fi
#2}}
\providecommand{\BIBdecl}{\relax}
\BIBdecl

\bibitem{brown2010universal}
E.~Brown, N.~Rodenberg, J.~Amend, A.~Mozeika, E.~Steltz, M.~R. Zakin,
  H.~Lipson, and H.~M. Jaeger, ``Universal robotic gripper based on the jamming
  of granular material,'' \emph{Proceedings of the National Academy of
  Sciences}, vol. 107, no.~44, pp. 18\,809--18\,814, 2010.

\bibitem{rus2015design}
D.~Rus and M.~T. Tolley, ``Design, fabrication and control of soft robots,''
  \emph{Nature}, vol. 521, no. 7553, p. 467, 2015.

\bibitem{cianchetti2013stiff}
M.~Cianchetti, T.~Ranzani, G.~Gerboni, I.~De~Falco, C.~Laschi, and
  A.~Menciassi, ``Stiff-flop surgical manipulator: mechanical design and
  experimental characterization of the single module,'' in \emph{2013 IEEE/RSJ
  International Conference on Intelligent Robots and Systems}.\hskip 1em plus
  0.5em minus 0.4em\relax IEEE, 2013, pp. 3576--3581.

\bibitem{manti2016stiffening}
M.~Manti, V.~Cacucciolo, and M.~Cianchetti, ``Stiffening in soft robotics: A
  review of the state of the art,'' \emph{IEEE Robotics \& Automation
  Magazine}, vol.~23, no.~3, pp. 93--106, 2016.

\bibitem{Wall2015}
V.~Wall, R.~Deimel, and O.~Brock, ``{Selective stiffening of soft actuators
  based on jamming},'' in \emph{2015 IEEE International Conference on Robotics
  and Automation (ICRA)}, 2015, pp. 252--257.

\bibitem{Amend2012}
J.~R. Amend, E.~Brown, N.~Rodenberg, H.~M. Jaeger, and H.~Lipson, ``{A positive
  pressure universal gripper based on the jamming of granular material},''
  \emph{IEEE Transactions on Robotics}, vol.~28, no.~2, pp. 341--350, 2012.

\bibitem{Jiang2012}
A.~Jiang, G.~Xynogalas, P.~Dasgupta, K.~Althoefer, and T.~Nanayakkara,
  ``{Design of a variable stiffness flexible manipulator with composite
  granular jamming and membrane coupling},'' in \emph{2012 IEEE/RSJ
  International Conference on Intelligent Robots and Systems}, 2012, pp.
  2922--2927.

\bibitem{steltz2009jsel}
E.~Steltz, A.~Mozeika, N.~Rodenberg, E.~Brown, and H.~M. Jaeger, ``Jsel:
  Jamming skin enabled locomotion,'' in \emph{2009 IEEE/RSJ International
  Conference on Intelligent Robots and Systems}.\hskip 1em plus 0.5em minus
  0.4em\relax IEEE, 2009, pp. 5672--5677.

\bibitem{HawkesScienceRobotics2017}
E.~W. Hawkes, L.~H. Blumenschein, J.~D. Greer, and A.~M. Okamura, ``A soft
  robot that navigates its environment through growth,'' \emph{Science
  Robotics}, vol.~2, no.~8, p. eaan3028, 2017.

\bibitem{comer1963deflections}
R.~Comer and S.~Levy, ``Deflections of an inflated circular-cylindrical
  cantilever beam,'' \emph{AIAA journal}, vol.~1, no.~7, pp. 1652--1655, 1963.

\bibitem{sanan2011physical}
S.~Sanan, M.~H. Ornstein, and C.~G. Atkeson, ``Physical human interaction for
  an inflatable manipulator,'' in \emph{2011 Annual International Conference of
  the IEEE Engineering in Medicine and Biology Society}.\hskip 1em plus 0.5em
  minus 0.4em\relax IEEE, 2011, pp. 7401--7404.

\bibitem{allen2003robotic}
M.~A. Allen, ``Robotic bridge maintenance system,'' Jan.~14 2003, uS Patent
  6,507,163.

\bibitem{voisembert2013design}
S.~Voisembert, N.~Mechbal, A.~Riwan, and A.~Aoussat, ``Design of a novel
  long-range inflatable robotic arm: Manufacturing and numerical evaluation of
  the joints and actuation,'' \emph{Journal of Mechanisms and Robotics},
  vol.~5, no.~4, p. 045001, 2013.

\bibitem{perrot2010long}
Y.~Perrot, L.~Gargiulo, M.~Houry, N.~Kammerer, D.~Keller, Y.~Measson,
  G.~Piolain, and A.~Verney, ``Long reach articulated robots for inspection and
  light interventions in hazardous environments, recent robotics research,
  tests campaigns and process,'' in \emph{CARPI 2010, 1st International
  Conference on Applied Robotics for the Power Industry}, 2010.

\bibitem{koren1991inflatable}
Y.~Koren and Y.~Weinstein, ``Inflatable structure,'' Nov.~19 1991, uS Patent
  5,065,640.

\bibitem{mishima2003development}
D.~Mishima, T.~Aoki, and S.~Hirose, ``Development of pneumatically controlled
  expandable arm for search in the environment with tight access,'' in
  \emph{Field and Service Robotics}.\hskip 1em plus 0.5em minus 0.4em\relax
  Springer, 2003, pp. 509--518.

\bibitem{tsukagoshi2011tip}
H.~Tsukagoshi, N.~Arai, I.~Kiryu, and A.~Kitagawa, ``Tip growing actuator with
  the hose-like structure aiming for inspection on narrow terrain.''
  \emph{IJAT}, vol.~5, no.~4, pp. 516--522, 2011.

\bibitem{blumenschein2018helical}
L.~H. Blumenschein, N.~S. Usevitch, B.~H. Do, E.~W. Hawkes, and A.~M. Okamura,
  ``Helical actuation on a soft inflated robot body,'' in \emph{2018 IEEE
  International Conference on Soft Robotics (RoboSoft)}.\hskip 1em plus 0.5em
  minus 0.4em\relax IEEE, 2018, pp. 245--252.

\bibitem{greer2018obstacle}
J.~D. Greer, L.~H. Blumenschein, A.~M. Okamura, and E.~W. Hawkes,
  ``Obstacle-aided navigation of a soft growing robot,'' in \emph{2018 IEEE
  International Conference on Robotics and Automation (ICRA)}.\hskip 1em plus
  0.5em minus 0.4em\relax IEEE, 2018, pp. 1--8.

\bibitem{coad2019retraction}
M.~M. Coad, R.~P. Thomasson, L.~H. Blumenschein, N.~S. Usevitch, E.~W. Hawkes,
  and A.~M. Okamura, ``Retraction of soft growing robots without buckling,''
  2019, (Submitted).

\bibitem{godaba2019payload}
H.~Godaba, F.~Putzu, T.~Abrar, J.~Konstantinova, and K.~Althoefer, ``Payload
  capabilities and operational limits of eversion robots,'' in \emph{Annual
  Conference Towards Autonomous Robotic Systems}.\hskip 1em plus 0.5em minus
  0.4em\relax Springer, 2019, pp. 383--394.

\bibitem{kim2013novel}
Y.-J. Kim, S.~Cheng, S.~Kim, and K.~Iagnemma, ``A novel layer jamming mechanism
  with tunable stiffness capability for minimally invasive surgery,''
  \emph{IEEE Transactions on Robotics}, vol.~29, no.~4, pp. 1031--1042, 2013.

\bibitem{Follmer2012}
\BIBentryALTinterwordspacing
S.~Follmer, D.~Leithinger, A.~Olwal, N.~Cheng, and H.~Ishii, ``{Jamming User
  Interfaces: Programmable Particle Stiffness and Sensing for Malleable and
  Shape-Changing Devices},'' \emph{Proceedings of the 25th annual ACM symposium
  on User interface software and technology - UIST '12}, pp. 519--528, 2012.
  [Online]. Available:
  \url{http://dl.acm.org/citation.cfm?doid=2380116.2380181}
\BIBentrySTDinterwordspacing

\bibitem{cheng2012design}
N.~G. Cheng, M.~B. Lobovsky, S.~J. Keating, A.~M. Setapen, K.~I. Gero, A.~E.
  Hosoi, and K.~D. Iagnemma, ``Design and analysis of a robust, low-cost,
  highly articulated manipulator enabled by jamming of granular media,'' in
  \emph{2012 IEEE International Conference on Robotics and Automation}.\hskip
  1em plus 0.5em minus 0.4em\relax IEEE, 2012, pp. 4328--4333.

\bibitem{Firouzeh2017}
A.~Firouzeh, M.~Salerno, and J.~Paik, ``{Stiffness Control with Shape Memory
  Polymer in Underactuated Robotic Origamis},'' \emph{IEEE Transactions on
  Robotics}, vol.~33, no.~4, pp. 765--777, 2017.

\bibitem{Alambeigi2016}
F.~Alambeigi, R.~Seifabadi, and M.~Armand, ``{A continuum manipulator with
  phase changing alloy},'' \emph{Proceedings - IEEE International Conference on
  Robotics and Automation}, vol. 2016-June, pp. 758--764, 2016.

\bibitem{liu2016interactive}
Y.~Liu, C.~Wang, H.~Tan, and M.~Wadee, ``The interactive bending wrinkling
  behaviour of inflated beams,'' \emph{Proceedings of the Royal Society A:
  Mathematical, Physical and Engineering Sciences}, vol. 472, no. 2193, p.
  20160504, 2016.

\end{thebibliography}

\end{document}